\documentclass{article}

\usepackage{microtype}
\usepackage{graphicx}
\usepackage{subcaption}
\usepackage{booktabs} 
\usepackage{enumitem}
\usepackage{hyperref}

\usepackage[accepted]{icml2026}

\usepackage{amsmath}
\usepackage{amssymb}
\usepackage{mathtools}
\usepackage{amsthm}
\usepackage{multirow}
\usepackage{array}      
\usepackage{booktabs}   
\usepackage{float}

\usepackage[capitalize,noabbrev]{cleveref}

\theoremstyle{plain}

\theoremstyle{definition}

\theoremstyle{remark}

\usepackage[textsize=tiny]{todonotes}

\icmltitlerunning{Submission and Formatting Instructions for ICML 2026}

\begin{document}

\twocolumn[
  \icmltitle{AnyBand-Diff: A Unified Remote Sensing Image \\ Generation and Band Repair Framework with Spectral Priors}



  \icmlsetsymbol{equal}{*}

  \begin{icmlauthorlist}
    \icmlauthor{Zuopeng Zhao}{cumt,erc,jiot}
    \icmlauthor{Ying Liu}{cumt}
    \icmlauthor{Xiaoyu Li}{cumt}
    \icmlauthor{Su Luo}{cumt}
    \icmlauthor{Lu Li}{cumt}
    \icmlauthor{Wenwen Liu}{cumt}
  \end{icmlauthorlist}
  \icmlaffiliation{cumt}{School of Computer Science and Technology / School of Artificial Intelligence, China University of Mining and Technology, Xuzhou, China}
  \icmlaffiliation{erc}{Mine Digitization Engineering Research Center of the Ministry of Education, Xuzhou, China}
  \icmlaffiliation{jiot}{Jiangsu Provincial Industrial Technology Engineering Center for Intelligent Sensing and Emergency IoT in Underground Space, Xuzhou, China}

  \icmlcorrespondingauthor{Ying Liu}{ts23170115p31@cumt.edu.cn}

  \icmlkeywords{Machine Learning, ICML}

  \vskip 0.3in
]



\printAffiliationsAndNotice{}  

\begin{abstract}
Existing diffusion models have made significant progress in generating realistic images. However, their direct adaptation to remote sensing imagery often disregards intrinsic physical laws. This oversight frequently leads to spectral distortion and radiometric inconsistency, severely limiting the scientific utility of generated data. To address this issue, this paper introduces AnyBand-Diff, a novel spectral-prior-guided diffusion framework tailored for robust spectral reconstruction. Specifically, we design a Masked Conditional Diffusion backbone integrated with a dual stochastic masking strategy, empowering the model to recover complete spectral information from arbitrary band subsets. Subsequently, to ensure radiometric fidelity, a Physics-Guided Sampling mechanism is proposed, leveraging gradients from a differentiable physical model to explicitly steer the denoising trajectory toward the manifold of physically plausible solutions. Furthermore, a Multi-Scale Physical Loss is formulated to enforce rigorous constraints across pixel, region, and global levels in a joint manner.  Extensive experiments confirm the effectiveness of AnyBand-Diff in generating reliable imagery and achieving accurate spectral reconstruction, contributing to the advancement of physics-aware generative methods for Earth observation.
\end{abstract}

\section{Introduction}

Remote sensing (RS) imagery stands as a cornerstone for a myriad of critical applications, ranging from environmental monitoring and disaster management to urban planning \cite{daras2023ambient} \cite{11242875} \cite{10107597}. In these domains, the capacity to accurately analyze and interpret data from satellite and airborne sensors is pivotal for informed decision-making \cite{de2025bird} \cite{10109830}. However, the efficacy of deep learning paradigms in this field is inherently predicated on the availability of large-scale, high-fidelity annotated datasets. Regrettably, the acquisition of such data remains a formidable challenge, plagued by prohibitive costs, atmospheric perturbations (e.g., cloud occlusion), and sensor-specific constraints \cite{hussain2022spatiotemporal, burke2021changing}. These impediments create a severe data scarcity bottleneck, substantially impeding the advancement of RS applications, particularly in resource-constrained regions or under adverse environmental conditions \cite{shirmard2022review} \cite{11223686}.

To mitigate these data constraints, Generative Artificial Intelligence (GenAI)—and specifically diffusion probabilistic models—has emerged as a paradigm-shifting solution, offering the capability to synthesize realistic imagery to alleviate data paucity \cite{zhang2017stackgan} \cite{11133727} \cite{11243913} \cite{esser2024scaling}. Diffusion models, in particular, have demonstrated remarkable potential in reconstructing high-quality visual content from minimal priors \cite{11299098} \cite{10440324} \cite{11071319}. Nevertheless, a fundamental challenge remains largely unaddressed: unlike natural imagery, which prioritizes perceptual plausibility, remote sensing data constitutes rigorous physical measurements of surface properties, such as radiance and reflectance across specific spectral bands \cite{yu2025guideline, dubovik2021grand}. Consequently, for generative models deployed in RS, visual fidelity alone is insufficient. It is imperative that these models maintain physical fidelity, ensuring that every generated pixel strictly adheres to radiometric consistency and faithfully reflects real-world spectral signatures.
\begin{figure*}[h]
\centering
\includegraphics[width=1\textwidth]{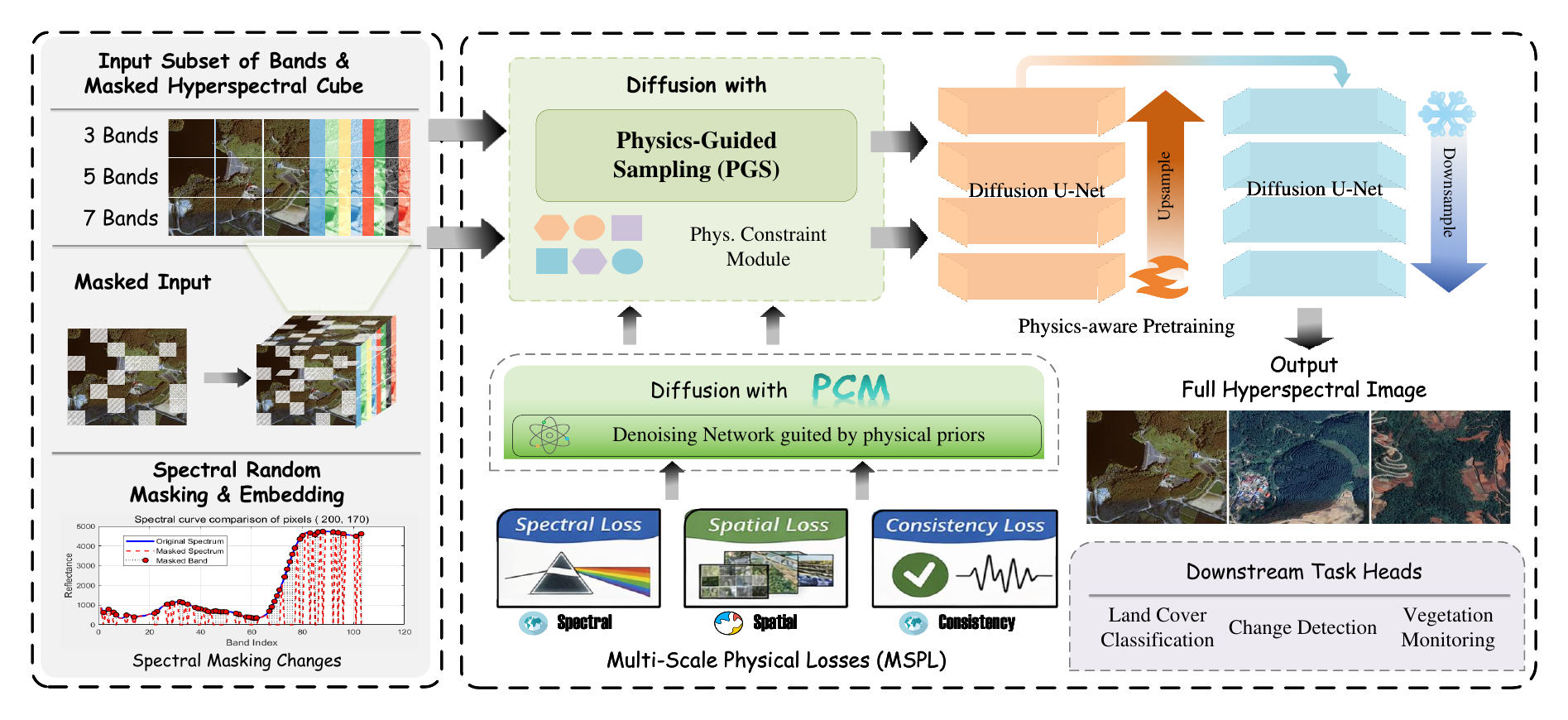}
\caption{The overall architecture of the AnyBand-Diff model. The input consists of a subset of spectral bands, which are randomly masked and embedded to simulate real-world sensor limitations or data loss. The model employs a PGS module to ensure spectral fidelity and physical consistency in the generated hyperspectral images. The diffusion process is carried out through Diffusion U-Net networks, with a pretraining step that incorporates physics-aware loss functions. The generated full hyperspectral image can be used for various downstream tasks such as land cover classification, change detection, and vegetation monitoring.}
\label{fig:architecture}
\end{figure*}
Despite their promise, existing generative frameworks suffer from critical deficiencies regarding physical consistency. A primary limitation is that these models often over-rely on perceptual texture synthesis, neglecting the underlying physical principles governing light-material interactions. This oversight frequently leads to spectral hallucinations, resulting in deviations in critical biophysical indices (e.g., NDVI) that are vital for quantitative analysis. A second major drawback is the rigidness of conditional generation mechanisms. Most prevalent models are trained on fixed input-output mappings (e.g., RGB-to-Multispectral), rendering them ill-suited for heterogeneous real-world scenarios where input data may stem from diverse sensors or suffer from arbitrary band loss \cite{goktepeecomapper}. This lack of flexibility severely curtails their applicability to incomplete or multi-source datasets. Furthermore, there exists a notable imbalance in multi-scale fidelity constraints. Conventional single-scale loss functions typically optimize for global image plausibility but fail to enforce physical realism across pixel, object, and scene levels—a hierarchical consistency that is indispensable for RS tasks requiring both fine-grained spectral precision and broad contextual coherence.

To bridge these gaps, we present \textbf{AnyBand-Diff}, a novel spectral-prior-guided diffusion framework tailored specifically for the physical rigors of remote sensing data. The core objective of AnyBand-Diff is to explicitly embed physical priors into the generative lifecycle, guaranteeing that the synthesized imagery is not only visually compelling but also physically valid. Our key insight lies in leveraging spectral physical principles to constrain the model at multiple stages: guiding the sampling trajectory, refining intermediate states, and optimizing the training objectives.

In summary, our main contributions are as follows:

\begin{itemize}
\item We propose a sampling strategy that leverages gradients from a differentiable physical model to steer the reverse diffusion process, ensuring spectral accuracy and physical validity.

\item We develop a backbone with dual stochastic masking, enabling robust full-spectrum reconstruction from arbitrary band subsets to effectively handle sensor-corrupted data.
  
\item We formulate a hierarchical loss that enforces rigorous constraints across pixel, region, and global levels, jointly guaranteeing radiometric fidelity and spatial coherence.
\end{itemize}

\section{Related Work}
\subsection{Remote sensing image generation}

The landscape of remote sensing (RS) image generation has witnessed a paradigm shift driven by deep generative models, evolving from Generative Adversarial Networks (GANs) to the more recent Denoising Diffusion Probabilistic Models (DDPMs) \cite{ho2020denoising}. Early approaches primarily leveraged GAN-based architectures to address data scarcity. Seminal works such as CycleGAN \cite{zhu2017unpaired} and Pix2PixHD \cite{wang2018high} have been extensively adapted for RS tasks, including super-resolution, cloud removal, and domain adaptation. While effective in enhancing spatial resolution and in-painting missing regions, GANs are inherently plagued by training instability (e.g., non-convergence) and mode collapse. More critically, in the context of spectral data, GANs often prioritize perceptual plausibility over radiometric accuracy. Although specific variants like SpecGAN \cite{9950553} attempted to incorporate spectral constraints, they still struggle to model the complex, high-dimensional joint distributions of multi-band data, frequently resulting in spectral distortion.
Diffusion models have emerged as a robust alternative, offering high-fidelity synthesis across diverse RS tasks. Recent advancements have focused on enhancing spatial and semantic consistency. MapGen-Diff \cite{tian2024mapgen} and Sui et al. \cite{sui2024denoising} effectively leverage denoising diffusion bridges, semantic priors, and adversarial learning to improve structural fidelity in map generation and super-resolution tasks. In multi-modal contexts, RSVQ-Diffusion \cite{gao2025rsvq} and CC-Diff \cite{11187367} integrate transformer architectures and spatial conditioning for controlled synthesis. Furthermore, diffusion frameworks have been adapted for spectral analysis; SpectralDiff \cite{chen2023spectraldiff} and DiffUCD \cite{zhang2023diffucd} exploit spectral-spatial features for hyperspectral classification and unsupervised change detection, respectively, while CRS-Diff \cite{10663449} targets reconstruction from partial inputs.
Despite these advancements, a critical gap remains: the trade-off between visual realism and physical fidelity. Most existing diffusion models for RS treat spectral bands merely as multi-channel RGB extensions, disregarding the underlying radiative transfer mechanisms. This "physics-agnostic" nature leads to generated data that, while visually convincing, lacks pixel-level radiometric consistency and inter-band spectral correlation. Furthermore, current methods often exhibit limited robustness when handling incomplete data (e.g., random band loss) or adapting across disparate sensors.
\subsection{Physics-Informed Generative Models}

Integrating domain-specific physical knowledge into deep learning frameworks—often termed Physics-Informed Neural Networks (PINNs) \cite{raissi2019physics}—has emerged as a pivotal direction in remote sensing. Unlike purely data-driven approaches, these models explicitly embed physical constraints, such as radiative transfer equations (RTE) or sensor-specific response functions, into the optimization landscape. This paradigm ensures that the synthesized outputs respect fundamental physical laws governing light-material interactions, thereby enhancing generalizability and scientific validity. In the context of Earth observation, recent foundation models have begun to explore this synergy. For instance, PhySwin \cite{tangphyswin} integrates physical priors—including spectral response functions and radiometric constraints—directly into a Swin Transformer architecture. By aligning the feature representation with the physical characteristics of multispectral sensors, PhySwin significantly improves the fidelity of downstream tasks such as land cover classification. Similarly, in the realm of adversarial generation, PhysicsGAN \cite{pan2020physics} attempts to enforce physical consistency by incorporating governing equations into the discriminator or loss functions. These efforts represent a crucial step towards bridging the gap between statistical learning and physical modeling.

However, the effective amalgamation of physical priors with modern generative diffusion models remains an open challenge. Existing physics-informed approaches often suffer from two main limitations: (1) \textbf{Over-simplified Constraints:} Many methods rely on linearized or simplified physical models to maintain computational tractability, which fails to capture the complex non-linearities of spectral mixing in real-world scenes. (2) \textbf{Optimization Conflict:} Naively adding physical loss terms often leads to optimization conflicts with the diffusion objective (noise prediction), resulting in generated samples that satisfy physical equations but lack perceptual realism or diversity. Distinct from prior arts, our proposed \textbf{AnyBand-Diff} introduces a novel mechanism that injects physical guidance directly into the \textit{sampling trajectory} rather than solely relying on static loss constraints. By utilizing gradients from a differentiable physical model to steer the denoising process, we achieve a harmonious balance between the generative flexibility of diffusion models and the rigorous precision of spectral physics.

\section{Method}

\subsection{Masked Conditional Diffusion (MCD)}

Let $\mathbf{X}_0 \in \mathbb{R}^{H \times W \times B}$ denote the ground-truth hyperspectral image. Conventional conditional models typically rely on a rigid, deterministic mapping from a fixed input (e.g., RGB) to $\mathbf{X}_0$. However, this paradigm is ill-equipped to handle the stochastic nature of real-world remote sensing scenarios, which are characterized by heterogeneous sensor configurations and sporadic data loss. To surmount this limitation, we introduce the \textbf{Masked Conditional Diffusion (MCD)} framework. Unlike standard approaches, MCD is underpinned by a \textbf{Dual Stochastic Masking (DSM)} strategy that dynamically simulates input uncertainty during training, thereby forcing the model to learn a robust recovery manifold from arbitrary partial observations.

The DSM process constructs the conditional input pair $(\mathbf{C}, \mathbf{M})$ through a two-stage degradation pipeline. First, to address sensor heterogeneity, we maintain a library of spectral response functions $\mathcal{R}$. In each training iteration, a response function $R$ is sampled from $\mathcal{R}$ and applied to $\mathbf{X}_0$ to generate observed data $\mathbf{X}_{\text{obs}}$. Second, to enforce robustness against missing bands, we apply a pixel-wise random dropout. The final validity mask $\mathbf{M}$ at location $(i,j)$ and band $b$ is derived via a Bernoulli process: $\mathbf{M}_{i,j,b} = \mathbf{M}_{\text{sens}, i,j,b} \cdot m$, where $m \sim \text{Bernoulli}(1 - p_{\text{drop}})$. The effective condition is then strictly defined as the sparse tensor $\mathbf{C} = \mathbf{X}_{\text{obs}} \odot \mathbf{M}$.

To effectively translate these sparse spectral constraints into the generative process, we propose a \textbf{Conditional Adaptive Modulation (CAM)} mechanism, moving beyond simple input concatenation. CAM employs a lightweight encoder $\mathcal{E}$ to map the sparse condition $\mathbf{C}$ and mask $\mathbf{M}$ into a global context vector $\mathbf{h} = \mathcal{E}(\mathbf{C}, \mathbf{M})$. This context vector is projected to regress layer-specific scale $\gamma^l$ and shift $\beta^l$ parameters, which modulate the intermediate feature maps $\mathbf{F}^l$ of the U-Net via a spatially-adaptive affine transformation:
\begin{equation}
\mathbf{F}^l_{\text{mod}} = \gamma^l(\mathbf{h}) \odot \text{Norm}(\mathbf{F}^l) + \beta^l(\mathbf{h}).
\end{equation}
This mechanism ensures that physical spectral cues are hierarchically injected into both shallow and deep representations. Finally, the network $\epsilon_\theta$ is optimized by minimizing the noise prediction error conditioned on this dynamic masking:
\begin{equation}
\mathcal{L}_{\text{MCD}} = \mathbb{E}_{\mathbf{X}_0, \mathbf{C}, \mathbf{M}, t, \epsilon} \left[ \| \epsilon - \epsilon_\theta(\mathbf{X}_t, t, \mathbf{C}, \mathbf{M}) \|_2^2 \right].
\end{equation}

\subsection{Physics-Guided Sampling (PGS)}

While diffusion models excel at perceptual synthesis, their standard inference trajectory relies exclusively on learned data distributions, often disregarding explicit physical constraints. This "physics-blind" generation can lead to spectral inconsistencies that violate fundamental radiometric laws. To rectify this, we propose \textbf{Physics-Guided Sampling (PGS)}, a test-time optimization strategy that actively steers the denoising trajectory toward the manifold of physically plausible solutions by leveraging gradients from differentiable physical operators.

\textbf{Mechanism.} 
In the reverse diffusion process at timestep $t$, the network predicts the noise $\epsilon_\theta(\mathbf{X}_t, t, \mathbf{C}, \mathbf{M})$, which implicitly estimates the clean image $\hat{\mathbf{X}}_0^{(t)}$ via Tweedie’s formula. PGS intervenes in this step by imposing a differentiable physical forward model $\mathcal{F}_{\text{phy}}(\cdot)$. This model encapsulates domain-specific laws, such as radiative transfer equations or spectral index calculations (e.g., NDVI, NDWI). We define a time-dependent physical consistency loss $\mathcal{L}_{\text{phy}}^{(t)}$ as:
\begin{equation}
\mathcal{L}_{\text{phy}}^{(t)} = \| \mathcal{F}_{\text{phy}}(\hat{\mathbf{X}}_0^{(t)}) - \mathbf{P}_{\text{target}} \|_2^2,
\end{equation}
where $\mathbf{P}_{\text{target}}$ represents the physical ground truth or heuristic constraints derived from the conditional input $\mathbf{C}$ (e.g., preserving spectral correlation matrices or enforcing valid reflectance ranges).

\textbf{Gradient Injection.} 
Unlike standard sampling which blindly follows the score function, PGS computes the gradient of the physical loss with respect to the noisy latent, $\mathbf{g}_t = \nabla_{\mathbf{X}_t} \mathcal{L}_{\text{phy}}^{(t)}$. This gradient serves as a correction signal, guiding the update step to minimize physical violations. The modified noise estimate $\tilde{\epsilon}_\theta$ is formulated as:
\begin{equation}
\tilde{\epsilon}_\theta = \epsilon_\theta(\mathbf{X}_t, t, \mathbf{C}, \mathbf{M}) - s \cdot \sqrt{1-\bar{\alpha}_t} \cdot \mathbf{g}_t,
\end{equation}
where $s$ is a guidance scale factor controlling the strength of the physical constraint, and $\sqrt{1-\bar{\alpha}_t}$ normalizes the gradient scale across noise levels. By replacing $\epsilon_\theta$ with $\tilde{\epsilon}_\theta$ in the sampling equation (e.g., DDIM), we effectively perform a gradient descent step in the physical solution space at each iteration.

\textbf{Advantages.} 
The core merit of PGS lies in its \textit{modularity} and \textit{model-agnosticism}. The physical operator $\mathcal{F}_{\text{phy}}$ acts as a plug-and-play module that can be tailored to specific tasks (e.g., hyperspectral unmixing or agricultural monitoring) without retraining the backbone model. This explicitly enforces hard physical constraints during the generative process, ensuring the final output is not only visually coherent but also scientifically valid.

\subsection{Multi-Scale Physical Loss (MSPL)}

To enforce rigorous physical validity across hierarchical spatial dimensions, we formulate a \textbf{Multi-Scale Physical Loss (MSPL)}. This objective jointly constrains the generative process at pixel, region, and global levels, defined as:
\begin{equation}
\mathcal{L}_{\text{MSPL}} = \lambda_{\text{px}}\mathcal{L}_{\text{pixel}} + \lambda_{\text{reg}}\mathcal{L}_{\text{region}} + \lambda_{\text{img}}\mathcal{L}_{\text{image}}
\end{equation}
In Eq. (5), the balancing hyperparameters \(\lambda_{\mathrm{px}}\), \(\lambda_{\mathrm{reg}}\), and \(\lambda_{\mathrm{img}}\) control the relative contributions of the pixel-level, region-level, and image-level physical consistency terms, respectively. In our implementation, these hyperparameters are set as follows: $\lambda_{\mathrm{px}} = 1.0,\quad \lambda_{\mathrm{reg}} = 0.5,\quad \lambda_{\mathrm{img}} = 0.2$.
These values were selected based on validation-set tuning. The pixel-level term (\(\lambda_{\mathrm{px}}\)) is assigned the largest weight, as preserving pixel-wise spectral correlation matrices is the primary objective of the band repair task. The region-level and image-level terms serve as complementary physical regularizers that enforce distributional consistency at larger spatial scales. Empirically, we observed that setting \(\lambda_{\mathrm{reg}}\) or \(\lambda_{\mathrm{img}}\) too high tends to over-constrain the reconstruction and slightly reduce spectral flexibility.
\begin{figure*}[ht]
    \centering
    \includegraphics[width=1\textwidth]{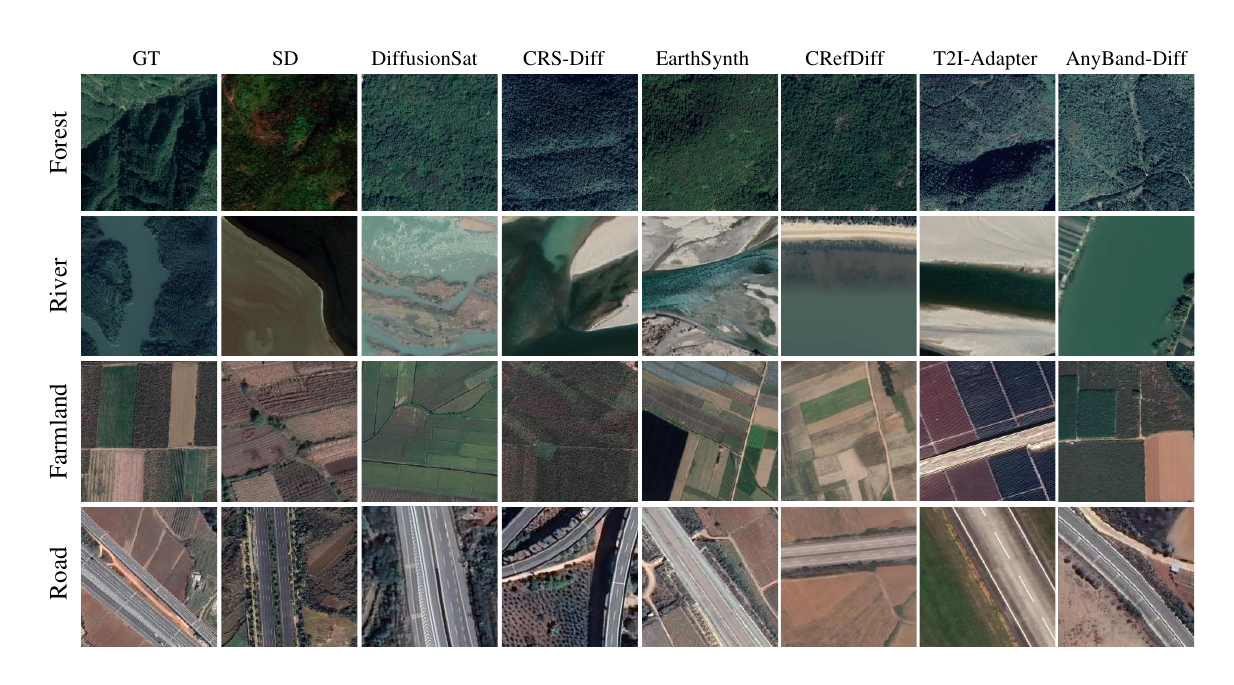}
    \caption{Comparison of remote sensing image generation results, showcasing the capability of generating RGB and panchromatic images.}
    \label{fig:prison_results}
\end{figure*}

\textbf{Pixel-Level: Spectral Correlation Constraint.} 
At the finest scale, it is crucial to preserve the intrinsic inter-band relationships (e.g., the specific reflectance curve of vegetation). We enforce this by aligning the spectral correlation matrix of the generated data $\hat{\mathbf{X}}_0$ with a prior matrix $\mathbf{S} \in \mathbb{R}^{B \times B}$, which encodes domain-specific spectral signatures (e.g., the negative correlation between Red and NIR bands). The loss is formulated as:
\begin{equation}
\mathcal{L}_{\text{pixel}} = \| \mathcal{C}(\hat{\mathbf{X}}_0) - \mathbf{S} \|_F^2,
\end{equation}
where $\mathcal{C}(\cdot)$ computes the Pearson correlation matrix over the spatial dimensions, and $\|\cdot\|_F$ denotes the Frobenius norm. This term prevents spectral distortion by locking the covariance structure of the synthesized bands.

\textbf{Region-Level: Distributional Physical Consistency.} 
At the local scale, we aim to maintain the statistical distribution of key biophysical indices (e.g., NDVI, NDWI) within local patches. Since standard histograms are non-differentiable, we employ a \textbf{Kernel Density Estimation (KDE)} based approach to approximate the distribution of physical indices. Let $\mathcal{F}_k^{\text{phy}}$ be the $k$-th differentiable index function, see Appendix a.1 for details. For a set of sampled patches $\{\mathbf{P}_n\}_{n=1}^N$, we minimize the Kullback-Leibler (KL) divergence between the estimated densities of real and generated features:
\begin{equation}
\small
\mathcal{L}_{\text{region}} = \frac{1}{N} \sum_{n=1}^{N} \sum_{k=1}^{K} D_{\text{KL}}\left( \text{KDE}(\mathcal{F}_k^{\text{phy}}(\mathbf{P}_n)) \parallel \text{KDE}(\mathcal{F}_k^{\text{phy}}(\hat{\mathbf{P}}_n)) \right).
\end{equation}
This ensures that the local statistical properties of land cover types are faithfully reproduced.

\textbf{Global-Level: Radiative Transfer Consistency.} 
At the global scale, we introduce a simplified, differentiable Radiative Transfer Model (RTM), denoted as $\Phi_{\text{RT}}$, to constrain the overall radiometric fidelity. $\Phi_{\text{RT}}$ acts as a physics-based autoencoder that maps surface reflectance to Top-of-Atmosphere (TOA) radiance and back, enforcing the generated image to lie on the manifold of valid physical measurements. The loss minimizes the reconstruction error in the physical domain:
\begin{equation}
\mathcal{L}_{\text{image}} = \| \Phi_{\text{RT}}(\hat{\mathbf{X}}_0) - \Phi_{\text{RT}}(\mathbf{X}_0) \|_2^2.
\end{equation}
By explicitly penalizing deviations from the RTM manifold, this term ensures global radiometric integrity and valid atmospheric interaction simulations.

\section{Experiments}

This section outlines a series of interconnected and progressively challenging experiments designed to systematically evaluate the efficiency and overall performance of AnyBand-Diff. The experimental framework is centered on the hyperspectral remote sensing image generation task and includes comprehensive comparisons against several state-of-the-art baseline methods. For general image generation and conditional control, we compare against the widely adopted Stable Diffusion (SD) model \cite{rombach2022high} and its prominent conditioning frameworks, ControlNet \cite{zhang2023adding}. To address remote sensing-specific image synthesis, we include the recent diffusion-based model DiffusionSat \cite{ICLR2024_16c3c941}. For cross-modal generation in the remote sensing domain, we incorporate EarthSynth \cite{pan2025earthsynthgeneratinginformativeearth} and Text2Earth \cite{liu2025text2earth}.  Furthermore, for hyperspectral and multi-band image synthesis, we consider specialized contemporaneous methods such as SpectralDiff \cite{chen2023spectraldiff}, SSDiff \cite{zhong2024ssdiff} ,CRS-Diff and HSI-Gene \cite{11180796}.

\subsection{Visual Quality Evaluation}

In this subsection, we rigorously evaluate the visual realism and spatial fidelity of the generated imagery. Our goal is to assess how well AnyBand-Diff reconstructs complex remote sensing scenes—encompassing intricate textures (e.g., urban structures) and natural transitions (e.g., vegetation boundaries)—compared to the state-of-the-art baselines.

We employ standard image quality metrics to benchmark performance: Peak Signal-to-Noise Ratio (PSNR) and Structural Similarity Index (SSIM) to quantify pixel-level and structural fidelity, respectively; and Root Mean Square Error (RMSE) to measure global reconstruction deviation. Additionally, Spectral Angle Mapper (SAM) is included to assess the color/spectral vector fidelity relative to the ground truth.
\begin{table}[h]
\centering
\small
\setlength{\tabcolsep}{3pt}
\caption{Quantitative Comparison of Visual Quality. \textbf{Bold} indicates best, \underline{underlined} indicates second best.}
\begin{tabular}{lcccc}
\toprule
\textbf{Method} & \textbf{PSNR} (dB) $\uparrow$ & \textbf{SSIM} $\uparrow$ & \textbf{SAM} $\downarrow$ & \textbf{RMSE} $\downarrow$ \\
\midrule
SD & 9.25 & 0.41 & 0.075 & 0.098 \\
ControlNet & 10.14 & 0.49 & 0.062 & 0.085 \\
DiffusionSat & 9.86 & 0.48 & 0.068 & 0.092 \\
SpectralDiff & 10.02 & 0.48 & 0.065 & 0.088 \\
SSDiff & 11.78 & 0.58 & 0.052 & 0.072 \\
EarthSynth & 14.48 & 0.64 & 0.046 & 0.065 \\
CRS-Diff & 15.07 & 0.69 & 0.034 & 0.048 \\
Text2Earth	& 14.92	& 0.67	& 0.038	& 0.052 \\
HSI-Gene	& \underline{15.83}	& \underline{0.71}	& \underline{0.031}	& \underline{0.044}\\
\midrule
\textbf{AnyBand-Diff} & \textbf{17.11} & \textbf{0.73} & \textbf{0.028} & \textbf{0.039} \\
\bottomrule
\end{tabular}
\label{tab:comparison_quality}
\end{table}

Table \ref{tab:comparison_quality} summarizes the quantitative results. \textbf{AnyBand-Diff} consistently outperforms all competing methods across all four metrics. Notably, our model achieves a PSNR of \textbf{17.11 dB} and an SSIM of \textbf{0.73}, significantly surpassing general-purpose generative models like Stable Diffusion and ControlNet. This superiority suggests that our architecture effectively mitigates the "hallucination" artifacts common in standard diffusion models, preserving the authentic structural layout of remote sensing objects. Furthermore, the lowest RMSE (\textbf{0.039}) and SAM (\textbf{0.028}) values indicate that AnyBand-Diff not only generates visually pleasing textures but also maintains high color fidelity and radiometric accuracy, closely aligning with the ground truth distribution.

\begin{table*}[h]
\centering
\small
\setlength{\tabcolsep}{4pt} 
\caption{Quantitative Results for Random Band Masking Reconstruction (Pavia University \& WashingtonDC Dataset). \textbf{Bold} indicates best performance.}
\begin{tabular}{lccccccccc}
\toprule
\multirow{2.5}{*}{\textbf{Method}} & \multicolumn{3}{c}{\textbf{10\% Masking}} & \multicolumn{3}{c}{\textbf{30\% Masking}} & \multicolumn{3}{c}{\textbf{50\% Masking}} \\
\cmidrule(lr){2-4} \cmidrule(lr){5-7} \cmidrule(lr){8-10}
& \textbf{PSNR} $\uparrow$ & \textbf{SSIM} $\uparrow$ & \textbf{RMSE} $\downarrow$ & \textbf{PSNR} $\uparrow$ & \textbf{SSIM} $\uparrow$ & \textbf{RMSE} $\downarrow$ & \textbf{PSNR} $\uparrow$ & \textbf{SSIM} $\uparrow$ & \textbf{RMSE} $\downarrow$ \\
\midrule
SD & 24.12 & 0.71 & 0.068 & 20.45 & 0.54 & 0.105 & 17.23 & 0.38 & 0.165 \\
ControlNet & 26.54 & 0.76 & 0.055 & 22.87 & 0.61 & 0.092 & 19.56 & 0.45 & 0.132 \\
DiffusionSat & 27.89 & 0.79 & 0.049 & 23.91 & 0.65 & 0.084 & 20.88 & 0.51 & 0.118 \\
SpectralDiff & 28.45 & 0.81 & 0.046 & 25.12 & 0.69 & 0.078 & 21.75 & 0.56 & 0.109 \\
SSDiff & 30.15 & 0.85 & 0.038 & 27.34 & 0.74 & 0.065 & 24.02 & 0.65 & 0.095 \\
EarthSynth & 31.88 & 0.88 & 0.032 & 28.56 & 0.78 & 0.058 & 25.45 & 0.70 & 0.082 \\
CRS-Diff & 32.14 & 0.89 & 0.030 & 29.10 & 0.80 & 0.054 & 24.12 & 0.72 & 0.078 \\
Text2Earth & 31.52 & 0.87	& 0.034	& 28.23	& 0.76	& 0.061	& 25.03	& 0.68	& 0.086 \\
HSI-Gene	& 33.06	& 0.90	& 0.027	& 30.15	& 0.82	& 0.049	& 26.87	& 0.74	& 0.069\\
\midrule
\textbf{AnyBand-Diff (Ours)} & \textbf{34.22} & \textbf{0.93} & \textbf{0.028} & \textbf{32.19} & \textbf{0.89} & \textbf{0.030} & \textbf{29.65} & \textbf{0.84} & \textbf{0.051} \\
\bottomrule
\end{tabular}
\label{tab:random_masking_results}
\end{table*}

Visual comparisons are presented in Figure \ref{fig:prison_results}. As illustrated, baseline methods (e.g., SD and ControlNet) often struggle with the specific characteristics of remote sensing data, occasionally producing blurred textures or geometrically distorted buildings. In contrast, AnyBand-Diff generates sharper edges and more coherent textures, particularly in challenging areas such as dense vegetation and water bodies. This qualitative advantage underscores our model's capability to synthesize high-fidelity imagery that is visually indistinguishable from real sensor data.
\begin{figure}[h]
    \centering
    \includegraphics[width=0.48\textwidth]{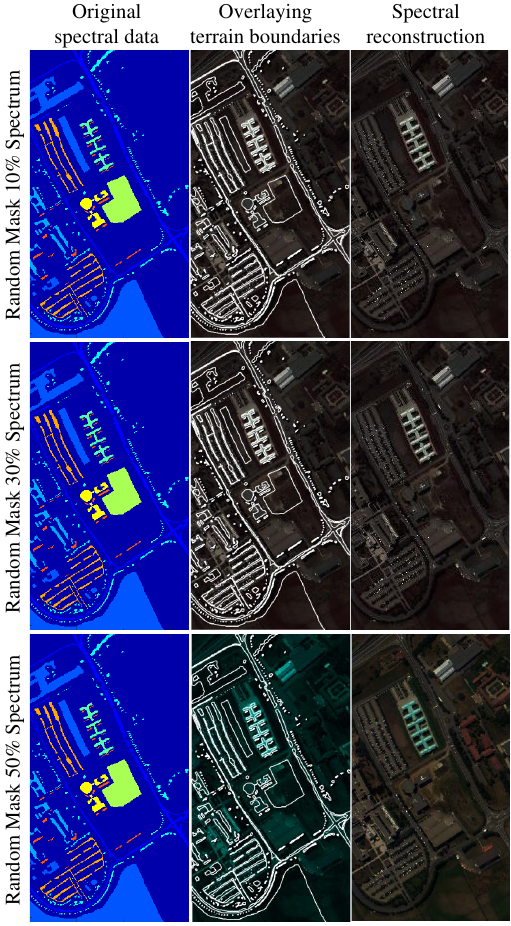}
    \caption{Quantitative comparison of AnyBand-Diff versus baseline methods on the hyperspectral image reconstruction task under different random band masking ratios (10\%, 30\%, 50\%).}
    \label{fig:random_masking_results1}
\end{figure}

\subsection{Evaluation of Spectral Reconstruction Robustness}

In this experiment, we rigorously evaluate the robustness of AnyBand-Diff in reconstructing hyperspectral data under varying degrees of spectral degradation. This setup simulates real-world challenges such as sensor malfunctions, cloud occlusion, or transmission errors, where random subsets of spectral bands are lost.

We introduce a \textit{Random Band Masking} protocol, where $10\%$, $30\%$, and $50\%$ of the spectral bands are stochastically dropped (zeroed out). The model is tasked with inferring the missing spectral information from the remaining partial observations. 
\textit{Note on Baselines:} It is important to acknowledge a structural disadvantage for general baselines (e.g., SD, ControlNet). These models typically treat multi-band inputs as generic multi-channel tensors, lacking explicit mechanisms to model inter-band spectral correlations. Consequently, they function as general in-painters rather than spectral reconstruction specialists. We include them to highlight the necessity of domain-specific spectral priors, which are central to our Masked Conditional Diffusion (MCD) framework.
\begin{figure}[h]
    \centering
    \includegraphics[width=0.5\textwidth]{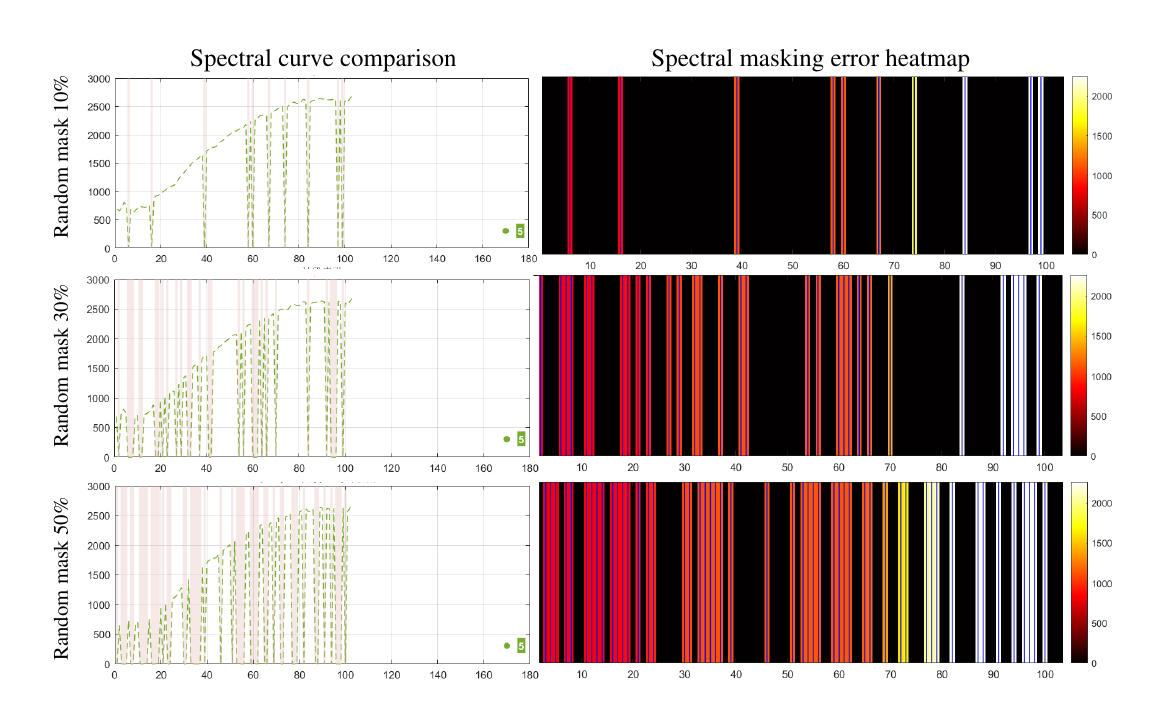}
    \caption{Schematic of band masking. In the spectral curve comparison, green points denote the masked bands. In the error heatmap, brighter colors indicate larger reconstruction errors.}
    \label{fig:random_masking_results2}
\end{figure}

The comparative results on the Pavia University dataset are presented in Table \ref{tab:random_masking_results}. \textbf{AnyBand-Diff exhibits exceptional robustness}, particularly in high-masking regimes.
\begin{itemize}
\item \textbf{Resistance to Degradation:} While generic models suffer catastrophic performance drops as masking increases (e.g., SD's PSNR plummets from 24.12 dB at 10\% to 17.23 dB at 50\%), AnyBand-Diff maintains high fidelity. Even with 50\% of bands missing, our model achieves a PSNR of \textbf{29.65 dB}, which is comparable to the performance of SSDiff with only 10\% masking (30.15 dB).
\item \textbf{Superiority Gap:} At the most challenging 50\% setting, our method outperforms the nearest competitor (HSI-Gene) by a substantial margin of \textbf{2.78 dB} in PSNR and \textbf{0.10} in SSIM. This widening gap underscores the effectiveness of our \textit{Dual Stochastic Masking} strategy, which forces the model to learn robust spectral dependencies during training.
\end{itemize}

In order to intuitively verify the spectral accuracy of our reconstruction, we give a detailed analysis in figure \ref{fig:random_masking_results1} and figure \ref{fig:random_masking_results2}.

\begin{table}[h]
\centering
\small
\setlength{\tabcolsep}{7pt}
\caption{Quantitative Comparison of Biophysical Indices Consistency. \textbf{Bold} indicates the best performance.}
\begin{tabular}{lcccc}
\toprule
\multirow{2.5}{*}{\textbf{Method}} & \multicolumn{2}{c}{\textbf{NDVI}} & \multicolumn{2}{c}{\textbf{NDWI}} \\
\cmidrule(lr){2-3} \cmidrule(lr){4-5}
 & \textbf{CC} $\uparrow$ & \textbf{RMSE} $\downarrow$ & \textbf{CC} $\uparrow$ & \textbf{RMSE} $\downarrow$ \\
\midrule
SD & 0.42 & 0.185 & 0.38 & 0.210 \\
ControlNet & 0.51 & 0.152 & 0.45 & 0.185 \\
DiffusionSat & 0.55 & 0.144 & 0.49 & 0.176 \\
SpectralDiff & 0.76 & 0.085 & 0.72 & 0.098 \\
SSDiff & 0.82 & 0.062 & 0.79 & 0.075 \\
EarthSynth & 0.85 & 0.045 & 0.81 & 0.052 \\
CRS-Diff & 0.88 & 0.038 & 0.84 & 0.046 \\
Text2Earth & 0.86 & 0.041 & 0.82 & 0.049 \\
HSI-Gene & 0.90 & 0.032 & 0.87 & 0.038 \\
\midrule
\textbf{AnyBand-Diff} & \textbf{0.94} & \textbf{0.015} & \textbf{0.92} & \textbf{0.022} \\
\bottomrule
\end{tabular}
\label{tab:biophysical_indices_comparison}
\end{table}

\begin{table*}[h]
\centering
\caption{Quantitative Results for Arbitrary Multiband Generation (Ablation Study)}
\setlength{\tabcolsep}{12pt} 
\begin{tabular}{c|ccc|c|c|c}
\hline
\multirow{2}{*}{\textbf{Method}} & \multicolumn{3}{c|}{\textbf{PSNR (dB)}} & \multirow{2}{*}{\textbf{SSIM}} & \multirow{2}{*}{\textbf{RMSE}} & \multirow{2}{*}{\textbf{SAM}} \\
\cline{2-4}
& \textbf{3 Bands} & \textbf{5 Bands} & \textbf{7 Bands} & & & \\
\hline
\textbf{Without Any Modules} & 28.75 & 30.11 & 31.01 & 0.82 & 0.050 & 0.038 \\
\textbf{Without MCD} & 33.18 & 34.01 & 34.89 & 0.91 & 0.032 & 0.029 \\
\textbf{Without PGS} & 32.10 & 33.48 & 34.15 & 0.89 & 0.036 & 0.031 \\
\textbf{Without MSPL} & 31.45 & 32.70 & 33.80 & 0.88 & 0.039 & 0.033 \\
\textbf{Without MCD and PGS} & 30.12 & 31.85 & 32.45 & 0.86 & 0.043 & 0.035 \\
\textbf{Full Model} & \textbf{34.25} & \textbf{35.42} & \textbf{36.01} & \textbf{0.93} & \textbf{0.028} & \textbf{0.025} \\
\hline
\end{tabular}
\label{tab:ablation_study_results}
\end{table*}

Figure \ref{fig:random_masking_results1} visualizes the spectral contour of representative pixels (such as vegetation, water, and city) together with the real value of the ground. According to the observation, the reconstructed spectral features closely track the original curve, and can accurately capture the key absorption features and reflection peaks even under 50\% masking. This confirms that anyband diff successfully learned the basic physical manifold of hyperspectral data.
Figure \ref{fig:random_masking_results2} shows the spatial distribution of spectral reconstruction error, in which the brighter the pixel, the greater the deviation. The error map is mainly dark, indicating that the residual noise in most scenes is minimal. It is worth noting that even in the heterogeneous transition region (for example, the boundary between buildings and roads), the error is still very low, indicating that the model can maintain spatial spectral consistency without introducing significant artifacts.

\subsection{Evaluation of Biophysical Consistency}

Beyond visual perception, the scientific utility of remote sensing imagery hinges on its ability to preserve accurate radiometric relationships. In this experiment, we assess the physical validity of the generated data by analyzing critical biophysical indices. Specifically, we focus on the Normalized Difference Vegetation Index (NDVI) and the Normalized Difference Water Index (NDWI), which are fundamental proxies for vegetation health and water body delineation, respectively.

We calculate NDVI and NDWI for both the ground truth (GT) and the generated imagery using the following standard spectral definitions:
\begin{equation}
\text{NDVI} = \frac{\text{NIR} - \text{Red}}{\text{NIR} + \text{Red}}, \quad \quad \text{NDWI} = \frac{\text{Green} - \text{NIR}}{\text{Green} + \text{NIR}}.
\end{equation}
To assess the physical consistency of the generated data, we compute the \textbf{Correlation Coefficient (CC)} between the generated index maps and the GT maps. CC measures the linear dependence and reflects the preservation of spatial patterns for these biophysical indices.
The results are reported in Table \ref{tab:biophysical_indices_comparison}. \textbf{AnyBand-Diff} demonstrates superior performance, achieving the highest correlation and lowest error across both indices.
\begin{itemize}
\item \textbf{Vegetation Fidelity (NDVI):} Our model achieves a CC of \textbf{0.94}, outperforming HSI-Gene (0.90) and significantly surpassing generic models like SD (0.42). This indicates that AnyBand-Diff accurately captures the "Red-Edge" characteristics and the steep reflectance contrast between Red and NIR bands, which are often smoothed out by standard generative models.
\item \textbf{Water Body Precision (NDWI):} Similarly, for NDWI, we achieve a CC of \textbf{0.92} and an RMSE of \textbf{0.022}. This proves the effectiveness of our \textit{Physics-Guided Sampling}, which steers the generation to respect specific spectral signatures of water bodies, avoiding the common issue where water is generated with incorrect spectral absorption features.
\end{itemize}

These results confirm that AnyBand-Diff does not merely hallucinate visually convincing textures but faithfully reconstructs the underlying physical properties of land cover, rendering the generated data valid for quantitative downstream applications.

\begin{table*}[h]
\centering
\caption{Computational efficiency comparison of different methods. All models are evaluated under the same input resolution (256$\times$256) on a single NVIDIA A100 GPU with FP16 precision. Inference uses DDIM with 50 sampling steps.}
\label{tab:efficiency}
\setlength{\tabcolsep}{15pt}
\begin{tabular}{lcccc}
\toprule
Method & Params (M) & FLOPs (G) & Latency (ms/img) & Sampling Steps \\
\midrule
SD & 860 & 58.2 & 2340 & 50 \\
ControlNet & 1,457 & 94.5 & 3120 & 50 \\
DiffusionSat & 312 & 28.6 & 1560 & 50 \\
EarthSynth & 289 & 26.3 & 1480 & 50 \\
CRS-Diff & 245 & 22.3 & 1280 & 50 \\
SpectralDiff & 186 & 18.5 & 1120 & 50 \\
SSDiff & 268 & 25.4 & 1460 & 50 \\
HSI-Gene & 412 & 35.7 & 1980 & 50 \\
AnyBand-Diff (w/o PGS) & 278 & 24.1 & 1350 & 50 \\
AnyBand-Diff (full) & 278 & 24.1 & 1420 & 50 \\
\bottomrule
\end{tabular}
\end{table*}

\subsection{Ablation Study for AnyBand-Diff}

In this experiment, we evaluate the performance of AnyBand-Diff by conducting an ablation study where different configurations of the model are tested.  These modules together enable the model to handle dynamic input uncertainties, enforce physical consistency during inference, and ensure consistency across different spatial scales during training.

To assess the contribution of each module, we progressively remove them in separate configurations. The results of this ablation study are presented in Table \ref{tab:ablation_study_results}, where we compare the performance of AnyBand-Diff under each configuration. The results in Table \ref{tab:ablation_study_results} show that AnyBand-Diff achieves the best performance across all band subsets (3, 5, and 7 bands) when all modules are present. Removing MCD results in a slight drop in performance, particularly in spectral consistency, as indicated by the higher SAM values. Without PGS, the model loses the ability to enforce physical consistency, which leads to an increase in RMSE and a decrease in PSNR. When MSPL is removed, the model's physical consistency across different spatial scales is compromised, resulting in worse performance, especially in the 7-band scenario. Removing both MCD and PGS further reduces performance, highlighting the importance of these components in ensuring both spectral fidelity and physical realism. Finally, the baseline model (without any modules) performs the worst, confirming the critical role of all modules in improving the overall performance. These results demonstrate the effectiveness of the key modules in AnyBand-Diff, showing that each component contributes significantly to the model's ability to generate high-quality and physically consistent multiband hyperspectral images.

\subsection{Efficiency Analysis}
To assess the practical utility of AnyBand-Diff, we conducted a comprehensive evaluation of its computational overhead against the main baseline methods. All models were measured under a unified experimental protocol with the same input resolution, batch size, hardware platform, and inference precision. For diffusion-based methods, inference was performed using DDIM sampling with 50 steps, consistent with the training noise schedule of \(T=1000\) timesteps. The results are reported in Table \ref{tab:efficiency}.

As shown in Table \ref{tab:efficiency}, AnyBand-Diff (full) contains 278 million parameters, with a computational cost of 24.1 G FLOPs and an end-to-end inference latency of 1420 ms per image. The proposed Physics-Guided Sampling (PGS) module introduces no additional parameters, as the physical model \(\mathcal{F}_{\text{phy}}\) remains frozen during inference. The slight latency increase compared to the w/o PGS version (70 ms, approximately 5\%) stems solely from the gradient computation required for physical guidance. Compared to HSI-Gene, our model uses 34\% fewer parameters and achieves 28\% lower latency, while delivering superior spectral reconstruction accuracy. These results demonstrate that AnyBand-Diff achieves a favorable trade-off between computational efficiency and generation quality, underscoring its potential for practical remote sensing applications.

\section{Conclusion}

In this work, we presented AnyBand-Diff, a pioneering framework that bridges the gap between generative visual synthesis and rigorous physical modeling in remote sensing. By synergizing a Masked Conditional Diffusion backbone with Physics-Guided Sampling, we have successfully addressed the long-standing visual-physical dilemma, enabling the generation of hyperspectral imagery that is not only perceptually indistinguishable from reality but also scientifically valid. Our extensive experiments demonstrate that AnyBand-Diff sets a new benchmark in spectral reconstruction, particularly in its robustness against severe data corruption (e.g., 50\% band loss) and its preservation of critical biophysical indices.
A pivotal insight of this study is that reliable AI-driven Earth observation hinges on explicit physical constraints, implemented either through multi-scale training objectives or gradient-guided inference. We have shown that treating spectral bands as mere image channels is insufficient; instead, modeling the underlying radiative manifold is key to unlocking the full potential of generative models for quantitative applications. Looking ahead, we aim to extend this physics-aware paradigm to spatiotemporal modeling for dynamic Earth monitoring and explore the potential of our generated data in enhancing the robustness of downstream quantitative analysis tasks.

\nocite{langley00}

\bibliography{example_paper}
\bibliographystyle{icml2026}

\newpage
\appendix
\onecolumn
\section{Appendix}
\subsection{Implementation Details of the Physical Model $\mathcal{F}_{\text{phy}}$}

The differentiable physical forward model, denoted as $\mathcal{F}_{\text{phy}}$, serves as the cornerstone of our Physics-Guided Sampling (PGS). To ensure computational efficiency and differentiability, we formulate $\mathcal{F}_{\text{phy}}$ using \textbf{closed-form differentiable operators} and a \textbf{lightweight neural emulator}, natively implemented in PyTorch. $\mathcal{F}_{\text{phy}}$ remains \textbf{frozen} during training to serve as an immutable physical anchor.

Specifically, $\mathcal{F}_{\text{phy}}$ comprises two primary components:

\textbf{1. Spectral Index Operators:}
We define differentiable index functions to enforce correct spectral topology for specific land covers. For a generated spectral vector $\hat{\mathbf{x}}$, the operators are:
\begin{equation}
    f_{\text{NDVI}}(\hat{\mathbf{x}}) = \frac{\hat{\mathbf{x}}_{\text{NIR}} - \hat{\mathbf{x}}_{\text{Red}} + \epsilon}{\hat{\mathbf{x}}_{\text{NIR}} + \hat{\mathbf{x}}_{\text{Red}} + \epsilon}, \quad f_{\text{NDWI}}(\hat{\mathbf{x}}) = \frac{\hat{\mathbf{x}}_{\text{Green}} - \hat{\mathbf{x}}_{\text{NIR}} + \epsilon}{\hat{\mathbf{x}}_{\text{Green}} + \hat{\mathbf{x}}_{\text{NIR}} + \epsilon}
\end{equation}
where $\epsilon=1e^{-6}$ ensures numerical stability.

\textbf{2. Radiative Consistency via Differentiable RTM Emulator:}
Directly integrating standard Radiative Transfer Models (RTMs) like PROSAIL or 6S is infeasible due to their non-differentiable nature and high computational cost. To overcome this, we employ a \textbf{Differentiable RTM Emulator}.
\begin{itemize}
    \item \textbf{Emulator Training:} Prior to the main training, we generated a synthetic dataset of 100,000 spectral pairs using the PROSAIL canopy radiative transfer model. The inputs are biophysical parameters (e.g., Leaf Area Index, Chlorophyll content), and the outputs are the corresponding canopy reflectance spectra. We then trained a lightweight Multi-Layer Perceptron (MLP) to serve as the surrogate model $\Phi_{\text{RTM}}$, mapping reflectance spectra $\hat{\mathbf{X}}$ back to their valid biophysical parameters $\mathbf{P}$.
    \item \textbf{Manifold Constraint:} During PGS, we enforce that the generated spectra $\hat{\mathbf{X}}$ must reside on the physical manifold defined by the RTM. We calculate the reconstruction loss by passing the estimated parameters back through the (frozen) emulator:
    \begin{equation}
        \mathcal{L}_{\text{RTM}} = \| \hat{\mathbf{X}} - \Phi_{\text{RTM}}^{-1}(\Phi_{\text{RTM}}(\hat{\mathbf{X}})) \|_2^2
    \end{equation}
    Alternatively, if the emulator is only forward ($\mathbf{P} \to \mathbf{X}$), we can minimize the distance between $\hat{\mathbf{X}}$ and the nearest physical neighbor found via gradient descent in the parameter space $\mathbf{P}$. In our efficient implementation, we use a pre-trained Autoencoder regularized by PROSAIL simulations, ensuring the decoder output strictly follows RTM laws.
    \item \textbf{Efficiency:} This neural surrogate approach is fully differentiable and incurs negligible computational overhead ($<5\%$ increase in inference time), enabling real-time physical guidance.
\end{itemize}

\subsection{Spectral Response Function Library ($\mathcal{R}$) and Alignment.}
We constructed a library $\mathcal{R}$ containing Spectral Response Functions (SRFs) from $K=15$ mainstream sensors (e.g., Sentinel-2, Landsat-8/9, EO-1 Hyperion). 
\textbf{Spectral Alignment:} Since different sensors operate on distinct spectral grids, we apply \textbf{Cubic Spline Spectral Resampling} to align the continuous SRF profiles with the discrete spectral bands of the input hyperspectral data before convolution. This ensures accurate simulation of sensor-specific observations regardless of the center wavelength mismatch.
During the \textit{Sensor Simulation Masking} phase, we uniformly sample a sensor $k$, resample its SRF $R_k$, and apply it to $\mathbf{X}_0$ to generate the physically valid conditional input.

\subsection{Dataset Description and Preparation}

To rigorously evaluate the generalization and spectral fidelity of AnyBand-Diff, we constructed a comprehensive benchmark consisting of both large-scale multispectral/RGB datasets and fine-grained hyperspectral datasets.

\paragraph{1) Large-Scale Multispectral \& RGB Datasets.}
These datasets are primarily used to evaluate the model's capability in generating visually realistic spatial structures and handling diverse land cover types.
\begin{itemize}
\item \textbf{Google Earth Imagery (RGB):} We curated a large-scale dataset of high-resolution optical remote sensing images from Google Earth, covering diverse scenes such as urban areas, farmlands, and forests. The dataset contains 50,000 images cropped to $512 \times 512$ pixels with a spatial resolution of 0.5m. This dataset serves as a foundational benchmark for visual quality evaluation.
\item \textbf{Landsat-8 OLI (Multispectral):} To test multi-band generation, we utilized the Landsat-8 Operational Land Imager (OLI) data. We selected 10,000 scenes with minimal cloud cover ($<5\%$). Each scene includes 9 spectral bands (Coastal to SWIR) with a spatial resolution of 30m. The data were radiometrically calibrated and atmospherically corrected to surface reflectance.
\end{itemize}

\paragraph{2) Fine-Grained Hyperspectral Datasets.}
These datasets are used for the core evaluation of spectral reconstruction accuracy and physical consistency (e.g., PGS and LMM modules).
\begin{itemize}
\item \textbf{Pavia University (PaviaU):} Acquired by the ROSIS sensor over Pavia, Italy. It consists of $610 \times 340$ pixels with 103 spectral bands ranging from 430 to 860 nm, with a geometric resolution of 1.3 meters.
\item \textbf{Washington DC (WDC):} Acquired by the HYDICE sensor over the Washington DC Mall. It contains $1208 \times 307$ pixels with 191 spectral bands (after removing water absorption bands) covering the 400–2400 nm range. This dataset is particularly challenging due to its complex urban materials and wide spectral range.
\end{itemize}

\paragraph{Data Preprocessing.}

\begin{itemize}
    \item \textbf{Normalization:} All spectral data were normalized to the range $[-1, 1]$ to match the dynamic range of the diffusion model input.
    \item \textbf{Patching:} Due to the large spatial dimensions, we employed a sliding window strategy.
\end{itemize}
\begin{itemize}
    \item For RGB/Landsat-8: Images were resized or cropped to $256 \times 256$.
    \item For Hyperspectral (PaviaU/WDC): We cropped patches of size $64 \times 64$ with a stride of 32 to maximize the number of training samples, resulting in approx. 12,000 patches for PaviaU and 8,000 for WDC.
\end{itemize}

\subsection{Implementation Details}
\textbf{Hardware \& Environment.}
All experiments were conducted on a high-performance computing cluster equipped with 4 $\times$ NVIDIA A100-PCIE-40GB GPUs. The implementation was based on PyTorch 2.1.0 with CUDA 11.8. We used Distributed Data Parallel (DDP) for multi-GPU training to accelerate convergence.

\textbf{Network Architecture.}
Our backbone is based on a modified U-Net architecture tailored for hyperspectral data:
\begin{itemize}
\item Input Channels: The first layer was modified to accept $B$ channels (e.g., 103 for PaviaU) instead of the standard 3 channels.
\item Model Size: The U-Net consists of 4 down-sampling and 4 up-sampling blocks, with channel multipliers of $[1, 2, 4, 8]$. The base channel width is set to 128.
\item Attention: Multi-head self-attention mechanisms are applied at the $16 \times 16$ and $8 \times 8$ resolutions.
\item Conditioning: The Conditional Adaptive Modulation (CAM) modules are inserted after each ResNet block to inject the sparse spectral features.
\end{itemize}

\textbf{Training Configuration.}
We trained the model for 500 epochs with a total batch size of 64 (16 per GPU).
\begin{itemize}
\item Optimizer: AdamW optimizer with $\beta_1=0.9, \beta_2=0.999$, and weight decay of $1e-4$.
\item Learning Rate: The initial learning rate was set to $1e-4$, utilizing a cosine annealing schedule with a linear warmup for the first 50 epochs.
\item Diffusion Setup: We employed a linear noise schedule ($\beta_{\text{start}}=1e-4, \beta_{\text{end}}=0.02$) with $T=1000$ diffusion steps.
\item Masking Strategy: During training, the random band dropout rate $p_{\text{drop}}$ was uniformly sampled from $[0.1, 0.7]$ to ensure robustness against various degrees of data loss.
\end{itemize}

\subsection{Baseline Reproduction}
To ensure a fair comparison, all baseline methods were retrained on the same datasets with consistent settings:
\begin{itemize}
\item Stable Diffusion (SD) \& ControlNet: We utilized the pre-trained SD v1.5 weights. Since SD is designed for RGB images, we employed a PCA-based dimensionality reduction to compress the hyperspectral data into 3 principal components for the latent space, and then projected back to the spectral space using a learned decoder.
\item DiffusionSat: We used the official implementation but fine-tuned the spectral encoder to match the band number of our datasets.
\item Training Budget: All baselines were trained until convergence, ensuring that performance differences stem from architectural designs rather than insufficient training.
\end{itemize}

\subsection{Limitations and Future Scope of the Physical Model}

While the proposed Physics-Guided Sampling (PGS) strategy significantly enhances spectral consistency, the current implementation of the physical model $\mathcal{F}_{\text{phy}}$ entails certain limitations that define its scope of applicability:
\begin{itemize}
\item Simplified Radiative Transfer Assumptions.
Our RTM emulator is primarily trained on the \textbf{PROSAIL} model, which assumes a turbid medium canopy. While effective for vegetation-dominated scenes (e.g., agriculture, forests), this approximation may be less accurate for complex 3D structures such as urban environments (e.g., high-rise buildings) or heterogeneous mixed pixels where geometric optical effects (e.g., shadowing, multiple scattering) dominate. For urban scenes, the model relies more on the statistical priors learned by the diffusion backbone rather than explicit radiative equations. Future work could integrate more diverse RTMs (e.g., \textbf{DART} for 3D urban scenes) into the emulator ensemble to broaden the physical validity.

\item Static Atmospheric Conditions.
The current physical constraints largely focus on surface reflectance properties, assuming that atmospheric correction has been implicitly handled or that the atmospheric state is relatively stable. We do not explicitly model spatially varying atmospheric parameters (e.g., aerosol optical depth, water vapor) within the $\mathcal{F}_{\text{phy}}$ loop. Consequently, in scenarios with heavy haze or thin clouds, the physical guidance might compete with the data-driven denoising process. Incorporating a dynamic atmospheric radiative transfer module (e.g., based on \textbf{6S}) would be a valuable extension to handle Top-of-Atmosphere (TOA) radiance generation directly.

\item Dependency on Emulator Precision.
The "differentiability" of our RTM is achieved via a neural surrogate (MLP). Although the emulator achieves high accuracy on synthetic validation sets, there exists an inherent \textbf{domain gap} between the synthetic training data (idealized parameters) and real-world spectral complexity. In rare cases, the emulator might provide suboptimal gradient directions if the generated spectrum falls outside the convex hull of the PROSAIL training distribution. We plan to mitigate this by employing \textbf{uncertainty-aware surrogates} (e.g., Bayesian Neural Networks) that can down-weight the physical loss when the emulator is uncertain.
\end{itemize}

\subsection{Cross-Validation of PGS and MSPL}
To evaluate the generalizability of our proposed Physics-Guided Sampling (PGS) and Multi-Scale Physical Loss (MSPL) modules, we integrated them into two representative baseline methods: SSDiff and CRS-Diff. The results are reported in Table \ref{tab:cross_validation}.

\begin{table}[h]
\centering
\caption{Cross-validation of PGS and MSPL on SSDiff and CRS-Diff under 50\% band masking (PSNR, dB).}
\label{tab:cross_validation}
\begin{tabular}{lcccc}
\toprule
Method & w/o PGS/MSPL & +PGS & +MSPL & +Both \\
\midrule
SSDiff & 24.02 & 25.18 & 24.89 & 26.03 \\
CRS-Diff & 24.12 & 25.34 & 25.01 & 26.21 \\
\bottomrule
\end{tabular}
\end{table}

As shown, both PGS and MSPL consistently improve the performance of existing methods when applied individually. The integration of both modules yields combined gains of 2.01 dB for SSDiff and 2.09 dB for CRS-Diff compared to their baseline versions. These results demonstrate that our proposed modules are not tied to any specific architecture but are generalizable components that can enhance the spectral reconstruction fidelity of existing diffusion-based remote sensing generation methods. This cross-validation further substantiates the broad applicability of our physics-guided paradigm.

\subsection{Ablation Study on the Guidance Scale \(s\) in Physics-Guided Sampling (PGS)}

We conducted an ablation study to analyze the effect of the guidance scale \(s\) in Eq. (4) on reconstruction performance. The experiments were performed on the 50\% random band masking task using the Pavia University dataset. We varied \(s\) from 0 to 2.0 and report the quantitative results in Table \ref{tab:guidance_scale}.

\begin{table}[htbp]
\centering
\caption{Ablation study on the guidance scale \(s\) for Physics-Guided Sampling (PGS).}
\label{tab:guidance_scale}
\begin{tabular}{cccc}
\toprule
\(s\) & PSNR (dB) \(\uparrow\) & SSIM \(\uparrow\) & SAM \(\downarrow\) \\
\midrule
0 (w/o PGS) & 27.68 & 0.81 & 0.058 \\
0.5 & 28.94 & 0.83 & 0.053 \\
1.0 & 29.65 & 0.84 & 0.051 \\
1.5 & 29.58 & 0.84 & 0.052 \\
2.0 & 28.91 & 0.82 & 0.056 \\
\bottomrule
\end{tabular}
\end{table}

Increasing \(s\) from 0 to 1.0 consistently improves reconstruction metrics, indicating that stronger physical guidance helps enforce spectral consistency during the reverse denoising process. The optimal performance is achieved at \(s = 1.0\), with improvements of 1.97 dB in PSNR and 0.007 in SAM compared to the baseline without PGS. However, when \(s\) becomes too large (e.g., 1.5–2.0), performance begins to deteriorate. This negative impact arises because an overly strong physical correction over-constrains the reverse denoising trajectory, partially overriding the learned diffusion prior and leading to over-correction. Consequently, the generated outputs become less faithful to the data distribution, manifesting as reduced reconstruction fidelity. Thus, the best performance is achieved at an intermediate \(s\) value, which strikes a balance between enforcing physical plausibility and preserving the generative flexibility of the diffusion model.

\end{document}